\documentclass[10pt,twocolumn,letterpaper]{article}

\usepackage{cvpr}              

\usepackage{graphicx}
\usepackage{graphics}
\usepackage{amsmath}
\usepackage{amssymb}
\usepackage{stfloats}
\usepackage{booktabs}
\usepackage[accsupp]{axessibility}

\usepackage[pagebackref,breaklinks,colorlinks]{hyperref}

\usepackage[capitalize]{cleveref}
\crefname{section}{Sec.}{Secs.}
\Crefname{section}{Section}{Sections}
\Crefname{table}{Table}{Tables}
\crefname{table}{Tab.}{Tabs.}

\def\cvprPaperID{5} 
\def\confName{CVPR}
\def\confYear{2023}

\begin{document}

\title{Digital Twin Tracking Dataset (DTTD): A New RGB+Depth 3D Dataset for Longer-Range Object Tracking Applications}

\author{Weiyu Feng$^*$, Seth Z. Zhao$^*$, Chuanyu Pan$^*$, Adam Chang, Yichen Chen, Zekun Wang, Allen Y. Yang\\
University of California, Berkeley\\
{\tt\small \{weiyu\_feng, 
sethzhao506,
chuanyu\_pan,
allenyang
\}@berkeley.edu}
}
\maketitle
\def\thefootnote{*}\footnotetext{Equal Contribution}
\def\thefootnote{\arabic{footnote}}

\begin{abstract}
   Digital twin is a problem of augmenting real objects with their digital counterparts. It can underpin a wide range of applications in augmented reality (AR), autonomy, and UI/UX. A critical component in a good digital-twin system is real-time, accurate 3D object tracking. Most existing works solve 3D object tracking through the lens of robotic grasping, employ older generations of depth sensors, and measure performance metrics that may not apply to other digital-twin applications such as in AR. In this work, we create a novel RGB-D dataset, called Digital Twin Tracking Dataset (DTTD), to enable further research of the problem and extend potential solutions towards longer ranges and mm localization accuracy. To reduce point cloud noise from the input source, we select the latest Microsoft Azure Kinect as the state-of-the-art time-of-flight (ToF) camera. In total, 103 scenes of 10 common off-the-shelf objects with rich textures are recorded, with each frame annotated with a per-pixel semantic segmentation and ground-truth object poses provided by a commercial motion capturing system. Through extensive experiments with model-level and dataset-level analysis, we demonstrate that DTTD can help researchers develop future object tracking methods and analyze new challenges. The dataset, data generation, annotation, and model evaluation pipeline are made publicly available as open source code at: \href{https://github.com/augcog/DTTDv1}{https://github.com/augcog/DTTDv1}.
\end{abstract}

\begin{table*}
    \centering
    \resizebox{\textwidth}{!}{
    \begin{tabular}{|c|c|c|c|c|c|c|c|c|c|c|c|}
    \hline
        dataset & data type & ToF sensor & texture & depth & occlusion & average distance & light variation & \# of frames & \# of scenes & \# of objects & \# of annotations\\
    \hline
        \text{FAT\cite{Tremblay_2018_CVPR_Workshops}} & synthetic & - & $\checkmark$ & $\checkmark$ & $\checkmark$ & - & $\checkmark$ & 60,000 & 3,075 & 21 & 205,359  \\
        \text{ShapeNet6D\cite{He_2022_CVPR}} & synthetic & - & $\checkmark$ & $\checkmark$ & $\checkmark$ & - & $\checkmark$ & 800,000 & - & 12,490 & - \\
        \text{StereoOBJ-1M\cite{liu2021stereobj1m}} & real & $\times$ & $\checkmark$ & $\times$ & $\checkmark$ & - & $\checkmark$ & 393,612 & 182 & 18 & 1,508,327 \\
        \text{TOD\cite{liu2020keypose}} & real & $\times$ & $\checkmark$ & $\checkmark$ & $\times$ & 0.65 & $\times$ & 64,000 & 10 & 20 & 64,000  \\
        \text{LINEMOD\cite{10.1007/978-3-642-37331-2_42}} & real & $\times$ & $\checkmark$ & $\checkmark$ & $\checkmark$ & 0.88 & $\times$ & 18,000 & 15 & 15 & 15,784  \\
        \text{YCB-Video\cite{xiang2018posecnn}} & real & $\times$ & $\checkmark$ & $\checkmark$ & $\checkmark$ & 0.85 & $\times$ & 133,936 & 92 & 21 & 613,917  \\
        \text{LabelFusion\cite{marion2018label}} & real & $\times$ & $\checkmark$ & $\checkmark$ & $\checkmark$ & 0.99 & $\checkmark$ & 352,000 & 138 & 12 & 1,000,000  \\
        \text{T-LESS\cite{hodan2017tless}} & real & $\checkmark$ & $\times$ & $\checkmark$ & $\checkmark$ & 0.77 & $\times$ & 47,762 & - & 30 & 47,762  \\
        \textbf{DTTD (Ours)} & real & $\checkmark$ & $\checkmark$ & $\checkmark$ & $\checkmark$ & 1.32 & $\checkmark$ & 55,691 & 103 & 10 & 136,226 \\
    \hline
    \end{tabular}}
    \caption{Comparison of DTTD dataset with other prior art. DTTD provides RGB-D data using Microsoft Azure Kinect ToF sensor for longer-range object pose estimation problems. DTTD also measures rich object texture in different lighting conditions. To make a fair comparison, we only present the average distance for real-world RGB-D datasets.}
    \label{tab:dataset_comparison}
\end{table*}

\section{Introduction}
\label{sec:intro}


Augmented reality (AR) is a growing area of research in both academia and industry. Two long-standing technical bottlenecks for AR applications are 3D localization and its solutions on mobile devices such as smartphones and wearable headsets. While localization using visual odometry in static environments is a relatively matured technology as evidenced by the number of available SLAM solutions \cite{ORBSLAM3_TRO, 8954208}, tracking individual objects in unknown or potentially cluttered environments is much more nascent. Potential solutions are also essential to underpin a category of augmented reality applications called digital twin. Examples include replacing paper manuals for machine maintenance with virtual 3D digital instructions and collaborative design of CAD models. 3D object tracking is an essential part of any good digital-twin solutions, aiming to estimate an object's rotation and translation relative to the camera with RGB or RGB-D input and worn by the user. 

Recent advances in precise 3D object tracking have been driven by deep neural networks (DNN), which rely on high-quality datasets to be trained effectively. Existing 3D object tracking datasets include synthetic datasets \cite{He_2022_CVPR, Tremblay_2018_CVPR_Workshops} and real-world datasets\cite{10.1007/978-3-642-37331-2_42, marion2018label, xiang2018posecnn, liu2020keypose, hodan2017tless, liu2021stereobj1m}, with many past demonstrations focusing on robot grasping tasks. Recent DNN models are widely trained on these datasets in an end-to-end manner to perform image semantic segmentation, object classification, and object pose estimation tasks\cite{ wang2019densefusion, He_2020_CVPR, He_2021_CVPR, Jiang_2022_CVPR, mo2022es6d}. By employing fusion techniques on aligned RGB images and depth maps, these approaches can be robust to varying lighting conditions and object occlusions. Many recent works are capable of performing 3D object tracking with sufficient precision in real-world applications \cite{wang2019densefusion}, demonstrating up to \emph{centimeter localization accuracy}, which may be sufficient for the use case of robot grasping.

However, these solutions would reveal a new set of challenges when applied to mobile AR applications. Real-world AR applications usually require accurate 3D object tracking at \emph{millimeter accuracy}, from observation distance up to one to two meters, and under various complex lighting conditions. Unfortunately, most existing real-world 3D object tracking datasets have been recorded with less than one meter distance and under a fixed lighting condition (see Table \ref{tab:dataset_comparison}).

Further compounding the above issues, these 3D object tracking datasets rely on older-generation depth sensors that are known to have high noise in depth estimation beyond one meter, such as using low-power stereo cameras or structured-light cameras. We believe these depth cameras have  hindered possible improvement of future algorithms to achieve accurate 3D pose estimation, especially for digital-twin AR applications. Finally, solving 3D pose estimation for AR applications has a certain level of expectation of real-time performance. However, when depth data in existing datasets are sparse given a 3D scene, running 3D pose estimation could be time consuming \cite{ku2018defense, He_2020_CVPR, He_2021_CVPR}.

In this paper, we introduce a new open-source dataset to as our first step to fill the above gaps, called Digital Twin Tracking Dataset (DTTD). As summarized in Table \ref{tab:dataset_comparison}, DTTD is designed to be a six degree-of-freedom (6-DoF) pose estimation dataset that aims to address the pose estimation problem for digital-twin applications. To this end, we select 10 common objects similar to \cite{doi:10.1177/0278364917700714, 7251504, 7254318}, purchasable online or at local grocery stores, placing them in an indoor room-scale environment where users using AR could perform interaction tasks with these objects from roughly 0.7 meter to 1.5 meters. Adopting Microsoft Azure Kinect time-of-flight (ToF) sensor, we collect aligned RGB and depth frames with the expectation of a large field of view, high image resolution ($1280\times 720$), and increased depth accuracy\cite{2020Senso..20.5104A, s22072469, s21020413}. The data is collected under different lighting conditions with different levels of object occlusions to simulate real-world digital-twin scenarios. We also create a novel data annotation pipeline that makes use of an OptiTrack\footnote{https://optitrack.com/} motion capture system to provide ground truth object pose annotations as well as per-pixel semantic segmentation in every frame. In the end, we evaluate the performance of the representative state-of-the-art pose estimation algorithms on DTTD, and analyze possible challenges that DTTD poses to the 6-DoF pose estimation task.

Our main contributions are summarized as follows:
\begin{enumerate}
    \item DTTD is a real-world, high-quality 3D object tracking dataset with 10 objects and 103 scenes designed for research in digital-twin AR use cases. This dataset provides image data with a state-of-the-art depth camera, providing high resolution, large field of view, and accurate dense depth map. The data are collected from a wide range of distances within 1.5 meters and with various lighting conditions.
    \item We open source a pipeline for generating 3D object tracking measurements using the OptiTrack motion capture system and also provide manually annotated and refined ground-truth scene labels. 
    \item Finally, we evaluate selected state-of-the-art methods on DTTD to demonstrate its utility and efficacy. We provide in-depth analysis from both model-level and dataset-level perspectives to illustrate new challenges in longer-range and wider lighting conditions.
\end{enumerate}

\begin{figure*}
        \centering
        \includegraphics[width=1.0\linewidth]{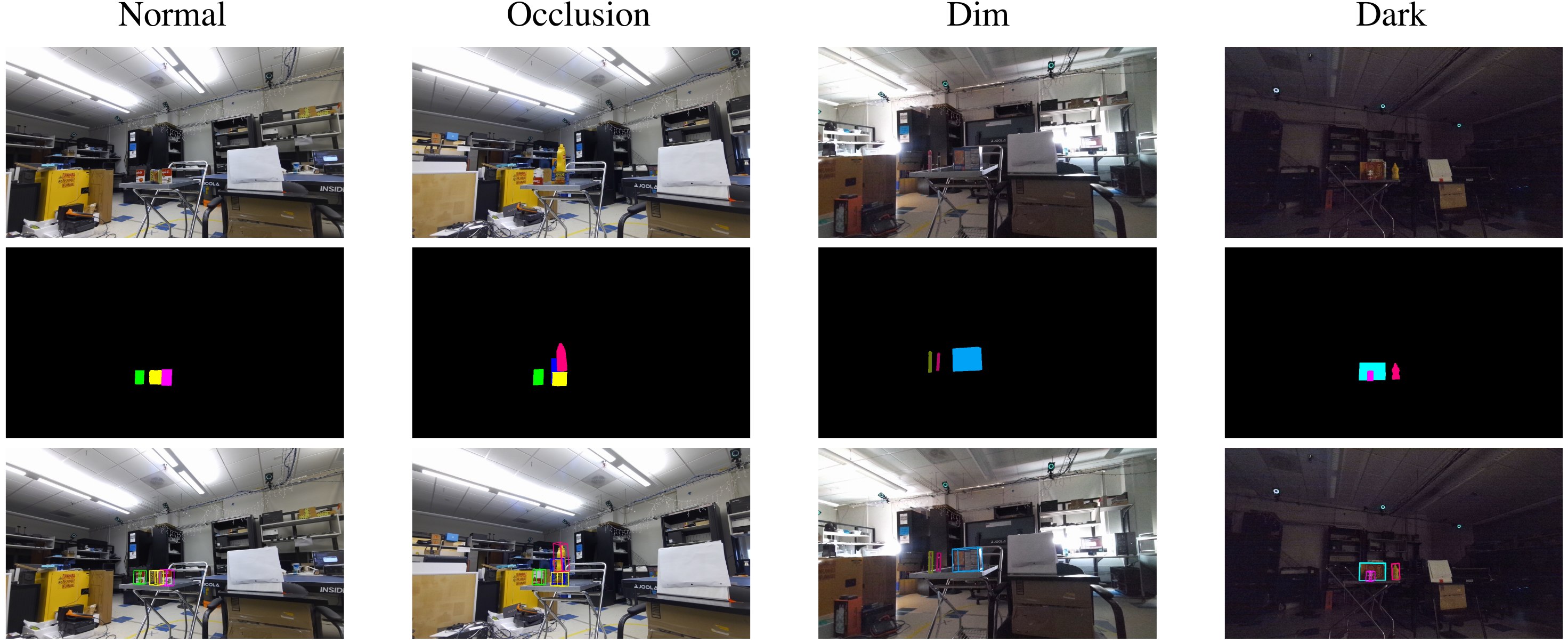}
    \caption{\textbf{Illustration of data samples.} DTTD dataset is collected under various real-world occlusion and lighting conditions. The first row shows scene images of four representative conditions (normal, occlusion, dim, dark). The second and third rows show the corresponding semantic segmentation labels and 3D bounding boxes, respectively. DTTD depth data have the same resolution as the scene images.}
    \label{fig:scenes}
\end{figure*}

\section{Related Work}
\subsection{3D Object Tracking Datasets} 
Existing object pose estimation algorithms are trained and evaluated primarily on a handful synthetic and real-world 3D object tracking datasets\cite{10.1007/978-3-642-37331-2_42, marion2018label, xiang2018posecnn, Tremblay_2018_CVPR_Workshops, He_2022_CVPR, liu2020keypose, hodan2017tless, liu2021stereobj1m}. Synthetic datasets use computer graphics rendering methods to generate photo-realistic images with various diverse shapes and appearances. For example,  FAT \cite{Tremblay_2018_CVPR_Workshops} renders 3D objects embedded in realistic background images to generate measurement with pose annotation. Shapenet6D \cite{He_2022_CVPR} generates images with augmented object shapes, texture, and background. Synthetic datasets typically require less human labor for annotation. However, models trained on synthetic datasets may have difficulties dealing with the domain-transfer problem when the systems are later applied in real world. 

In comparison, real-world datasets collect frames from real scenes and then employ manual annotation to label ground-truth 3D poses. Datasets such as YCB-Video\cite{xiang2018posecnn}, LINEMOD\cite{10.1007/978-3-642-37331-2_42}, StereoOBJ-1M\cite{liu2021stereobj1m}, and TOD\cite{liu2020keypose} utilize depth-from-stereo sensors to collect real 3D data. While previous generations of stereo sensors primarily target use cases within one-arm distance (i.e., about 0.7 meter), the depth accuracy would degrade very rapidly as the range increases \cite{Haggag2013MeasuringDA}. Furthermore, stereo sensors also suffer from holes in the depth map when stereo matching fails, requiring hole-filling algorithms\cite{ku2018defense} to preprocess the data and hence sacrifice the solution speed.

TLess\cite{hodan2017tless} is another option that collects data via time-of-flight (ToF) sensors, which improves the quality of depth data in room scale. Our DTTD dataset also uses state-of-the-art ToF camera, namely, Microsoft Azure Kinect, to capture meter-scale RGB-D data, and uses the OptiTrack system to collect the ground-truth camera position.

Regarding the pose annotation pipeline, most existing RGB-D datasets manually create their annotation by fitting 3D mesh models to 3D point cloud \cite{10.1007/978-3-642-37331-2_42, xiang2018posecnn, hodan2017tless, wang2019normalized}. In order to generate large quantities of labeled frames, annotation errors up to a centimeter are tolerated. For example, LabelFusion\cite{marion2018label} combines human keypoint labeling with matching algorithms to generate annotation. However, without per-frame manual refinement, centimeter-scale error may frequently occur, causing insufficient accuracy for tracking tasks in AR and digital-twin applications. DTTD relies on a combination of manual and automated refinement on each frame to improve the annotation quality compared to the existing works. More specifically, DTTD maintains accurate camera positions using the Optitrack system to keep objects' poses consistent among frames, thus reducing the annotation workload.

\subsection{6 DoF Object Pose Estimation} 
Most data-driven methods for object pose estimation take RGB\cite{labbe2020, peng2019pvnet, xiang2018posecnn, zakharov2019} or RGB-D images\cite{He_2021_CVPR, He_2020_CVPR, wang2019densefusion, mo2022es6d, Jiang_2022_CVPR} as input. Xiang et al.\cite{xiang2018posecnn} propose a convolutional neural network for 3D pose estimation problem. To fuse different data modalities more efficiently, Wang et al.\cite{wang2019densefusion} propose a network architecture to extract and combine pixel-wise dense feature embedding for both RGB and depth sources. Due to the architecture simplicity, this method reaches high efficiency on predicting object poses. More recent works\cite{He_2020_CVPR, He_2021_CVPR, He_2022_CVPR} improve the performance with more elaborate network architectures. For example, He et al.\cite{He_2021_CVPR} propose an improved bidirectional fusion network for key-point matching, and achieve high accuracy on YCB-Video\cite{xiang2018posecnn} and LINEMOD\cite{10.1007/978-3-642-37331-2_42} benchmarks. However, these methods are less efficient due to the complicated hybrid network structure and processing stages. For symmetric objects, Mo et al.\cite{mo2022es6d} propose a symmetry-invariant pose distance metric to avoid local minima issues. Jiang et al.\cite{Jiang_2022_CVPR} propose an L1-regularization loss, called abc loss, which improves the pose estimation accuracy for non-symmetric objects. Our work will focus on RGB-D data and evaluate some of the recent methods on DTTD to create a baseline benchmark.


\section{DTTD Dataset} \label{sec:DTTD}

DTTD collects RGB images and depth images from 10 rigid objects together with their corresponding textured 3D models. In total, there are 103 scenes that each includes one or more of such objects with different orientations. The dataset provides ground-truth labels about 3D object poses and per-pixel semantic segmentation. Camera specifications, pinhole camera projection matrices, and distortion coefficients are also provided in the dataset. The detailed dataset protocol is described below.

\subsection{Scenes}

A scene in DTTD consists of a static assortment of objects placed on a flat surface. Each scene is captured by a single RGB-D camera whose movement is tracked by a motion capture system. The camera movement is standardized as half-circle revolutions around the object(s), exposing different faces of the object(s) and varying levels of occlusion. Each scene contains up to five objects. As shown in Figure \ref{fig:scenes}, we purposefully collect scenes at multiple times of the day with different lighting conditions. The metadata for each frame contains labeled ground-truth object poses and camera intrinsic and distortion parameters. Users of DTTD may reproject the depth map into 3D point cloud using the camera parameters.

\subsection{Camera}

RGB images and corresponding depth maps are collected from a Microsoft Azure Kinect depth camera. During manual data collection, the camera is attached to a cart and pushed smoothly around the scene, minimizing the shaking from the handheld operation. We then retrieve aligned RGB and depth frames from Azure Kinect at $1280\times 720$ resolution and 30 frames per second (fps), with some tolerance of latency. The frames are aligned and synchronized and retrieved through the Azure Kinect SDK. The Azure Kinect SDK also provides radial and tangential distortion coefficients that we store in addition to the frames. 

\begin{figure*}
        \centering
        \includegraphics[width=1.0\linewidth]{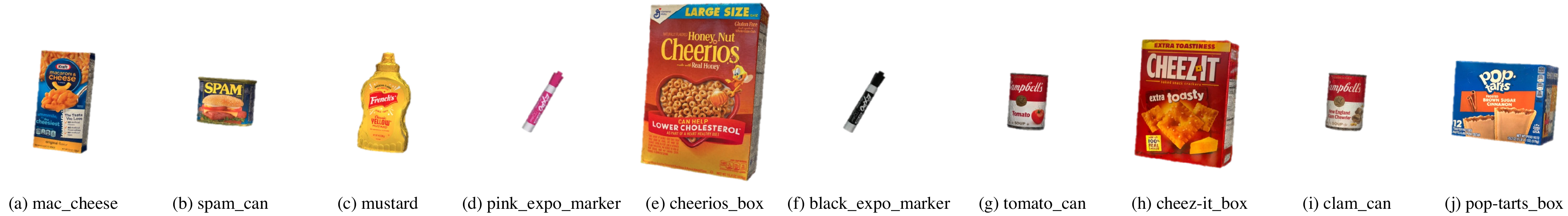}
    \caption{\textbf{Illustration of 3D CAD models of the 10 objects in DTTD.} Notice that \{d,f\} and \{g,i\} are two pairs of objects that are geometrically identical but have different color texture.
}
    \label{fig:objects}
\end{figure*}

\subsection{Object Models}

\begin{figure}
        \centering
        \includegraphics[width=1.0\linewidth]{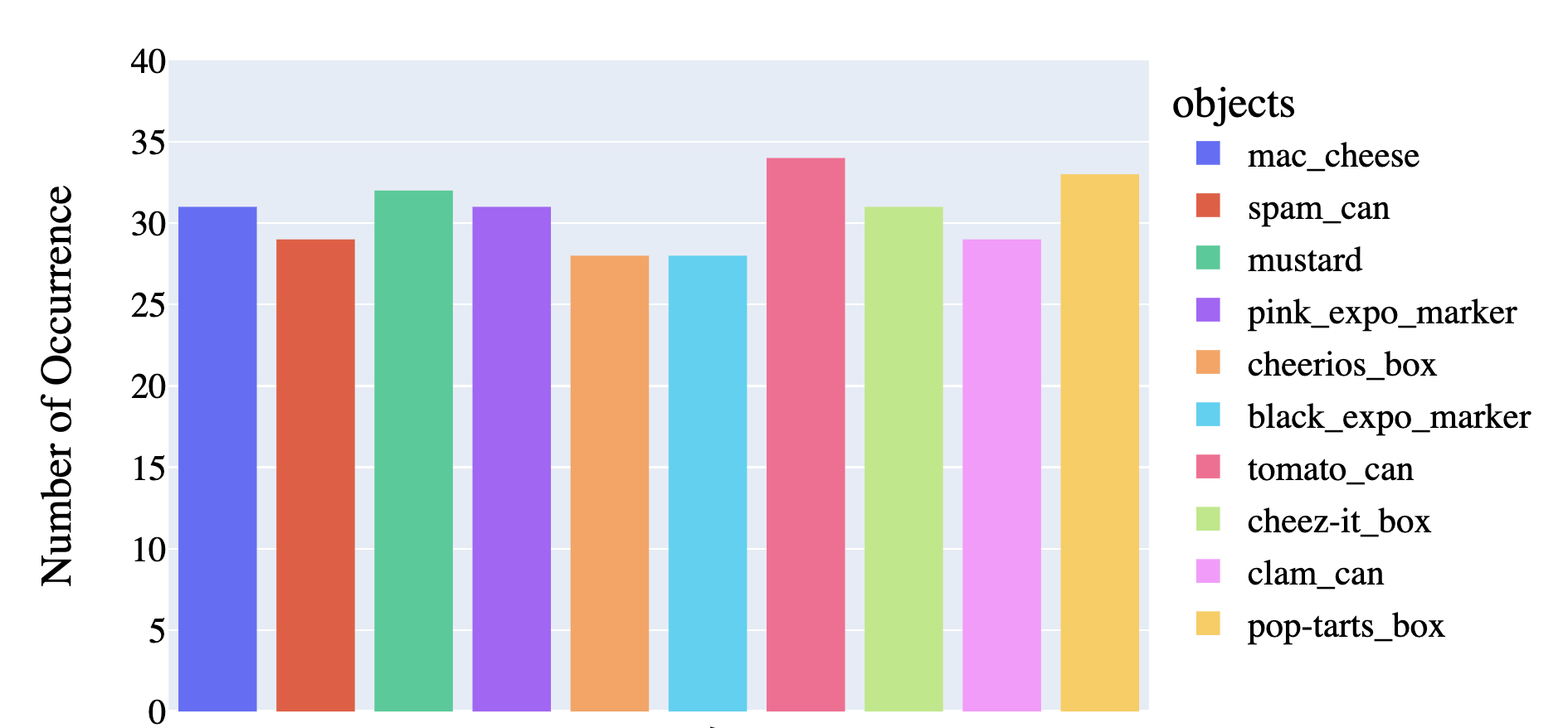}
    \caption{Number of scenes where objects occur in DTTD.}
    \label{fig:objects_occurence}
\end{figure}

DTTD provides textured 3D models for the 10 DTTD objects. These models are scanned using the Polycam app \footnote{https://poly.cam/} on an iPhone 12 Pro with its camera and LIDAR sensors. For model refinement, we use Blender \footnote{https://www.blender.org/} to fix surface holes and texture pixels that are scanned incorrectly. To make the collected scenes compatible with other existing datasets, part of the then DTTD objects overlap with the YCB-Video dataset \cite{xiang2018posecnn}. We use these 3D models to perform annotation of our data as well as generate semantic segmentation. Figure \ref{fig:objects} shows the collected models. Figure \ref{fig:objects_occurence} shows the distribution of the occurrence of the objects appearing in different scenes. The distribution is nearly uniform.

\subsection{Pose Annotation}

We have created a novel data annotation pipeline to annotate and refine ground-truth poses for objects in the scene. The pipeline first reprojects camera frames into corresponding 3D point cloud and aligns virtual models as the initial ground-truth pose. Then a combination of \emph{iterative closest points} (ICP)\cite{121791} registration methods and other refinement techniques are used to refine both the camera and object poses. Finally, a global pose refinement step is performed. By reviewing the annotation results of each frame and each object manually, investigators can ensure high-quality labels in the dataset. Annotating one scene can take as little as five minutes, depending on the quality of the automatic refinement. In order to generate the per-pixel semantic segmentation, the pipeline can further project the virtual models onto the 2D camera plane using the camera intrinsic and distortion parameters given the virtual models' known poses.

\subsection{Data Synthesizer and Scene Augmentation} \label{sec:synthetic-data}

The open-source DTTD code has the ability to further augment the dataset with synthetic images for training sophisticated deep learning models. We build our own data synthesizer using the Open3D Visualization platform \cite{Zhou2018}. Users are able to import custom CAD models, cameras, and backgrounds into the data synthesizer. The virtual cameras in simulation capture five kinds of data, including RGB images, depth images, semantic segmentation, object rotation, and translation. We also add Random Movement Component and Random Rotation Component to objects to create diverse synthetic scenes. As part of our data generation pipeline, we also provide the ability to render synthetic labeled data selected from the original DTTD object models.

\section{Data Generation Pipeline} \label{sec:methodology}

\subsection{Collection Setup}

\begin{figure}
        \centering
        \includegraphics[width=0.93\linewidth]{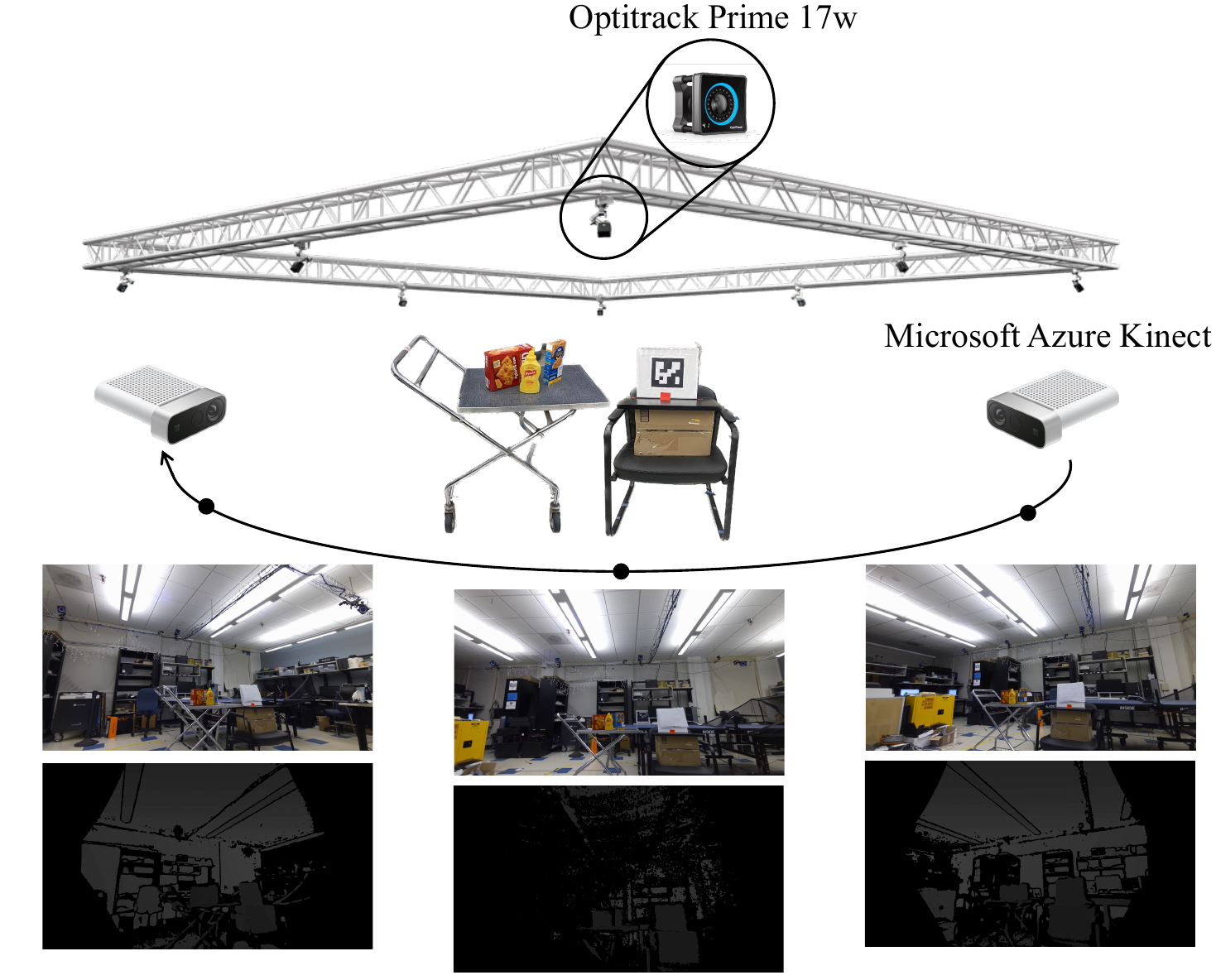}
    \caption{\textbf{DTTD setup.} The room is outfitted with 10 Prime 17W OptiTrack cameras to establish ground-truth 6 DoF orientation in real-time. With objects resting on a flat support platform, OptiTrack tracks the camera movement by OptiTrack markers glued onto the camera body. An ARUCO marker is placed in the scene to calculate the offset between the camera coordinate frame from where the objects are measured and the OptiTrack coordinate frame from where the camera movement is measured.}
    \label{fig:collect_data}
\end{figure}

We collect the data in a single room outfitted with an OptiTrack motion capture system with 10 Prime 17W cameras to cover the room space. Calibration of the system using OptiTrack's Motive software reports a mean 3D reprojection error of 0.8mm. We calibrate the depth camera to the OptiTrack system using an ARUCO marker in the center of the room. The camera is instrumented with six OptiTrack markers to define a rigid body. Then the single rigid body can be tracked by the OptiTrack software. The OptiTrack software returns a set of translation and rotation parameters for each rigid body in each frame during recording. Figure \ref{fig:collect_data} illustrates the setup of a standard scene. 

\subsection{Sensor Timestamp Synchronization}
The presence of multiple sensors during the data collection procedure requires reasonable time synchronization, although the synchronization does not need to be at high frame rates.

More specifically, the Azure Kinect camera captures frames at 30 fps, with some error due to system latency. The OptiTrack system is set up to output poses at 60 Hz. The two are then synchronized by matching computed poses from camera frames computed using the ARUCO marker with OptiTrack poses. We optimize the matching by minimizing the mean squared error. We use linear interpolation to estimate sub-interval poses from the OptiTrack. We also smooth the OptiTrack pose estimation jittering using a Kalman filter with a constant velocity model. This method generally yields good data alignment.

\subsection{Extrinsic Calibration}

One important step of our data generation pipeline is to ensure the OptiTrack is properly tracking the coordinate frame of the camera sensor itself, which we denote as $g_{oc}$ (transform from the camera sensor to OptiTrack). The OptiTrack is directly tracking the coordinate transform of the rigid body defined by its markers on the camera. This coordinate transform has a translation and rotation frame with respect to the actual camera sensor. We refer to this coordinate frame as $g_{ov}$ (transform from virtual camera to OptiTrack), or the virtual camera pose.

We refer to the extrinsic calibration as calculating the transformation between the OptiTrack's virtual camera rigid body coordinate system and the camera sensor's own coordinate system, or $g_{cv}$ (transform from virtual camera to camera sensor). In order to solve this extrinsic transformation, we first measure the position and rotation of an ARUCO marker in the OptiTrack world coordinate system, $g_{oa}$ (transform from ARUCO marker to OptiTrack). We then capture a sequence of frames with the camera observing the ARUCO marker and the OptiTrack observing the camera. Using the known pose of the ARUCO marker in the camera sensor $g_{ca}$ (transform from camera sensor to ARUCO marker), we can solve for the pose of the camera sensor in the OptiTrack world coordinate system, $g_{oc}$ (transform from the camera sensor to OptiTrack):
\begin{equation}
g_{oc} = g_{oa} \cdot g_{ca}^{-1}
\end{equation}

Now that we have a virtual camera pose $g_{ov}$ and a camera sensor pose $g_{oc}$, both in OptiTrack world coordinates, we can solve for an extrinsic transform for each frame and save the average extrinsic:
\begin{equation}
g_{cv} = g_{oc}^{-1}\cdot g_{ov}
\end{equation}
We perform averaging over the translation using the arithmetic mean and the rotation in the quaternion space.

Once this extrinsic transformation has been computed, as long as the OptiTrack system remains running, we can now directly track the pose of the camera sensor itself through OptiTrack's real-time tracking of the virtual camera rigid body, which significantly expands our ability in DTTD to track long-duration, large-transformation 3D motions while keeping precision of the pose labeling automatically.

\subsection{Annotation Pipeline}

\textbf{Initial Pose Annotation.} Our initial pose annotation is performed in a 3D environment using Open3D visualization framework \cite{Zhou2018}. We reproject RGB and depth maps into point cloud using camera intrinsic and distortion coefficients. We then place virtual models into the scene using ICP initialized by a rough user approximation of the center of each object in the scene. Annotators can then individually adjust position and rotation of virtual models as well as perform incremental algorithmic alignment until virtual models are fully aligned with the point cloud. Once the annotated poses for the objects are satisfactory, they are saved and used to transform the objects into every subsequent camera frame in the sequence. 

\textbf{Scene Pose Refinement.} After transforming each virtual object into every frame, we perform a frame-wide global pose refinement using ICP to refine camera poses. Finally, we have another manual refinement phase, where annotators can review each frame in 2D to ensure good alignment. The environment allows the annotator to adjust the camera translation and rotation in a very similar manner as in the object adjustments during the annotation phase.

\section{Experiments}
\subsection{Benchmark and Evaluation}
\textbf{Train/Test Split.} DTTD provides a recommended train/test split as follows. The training set contains 8726 keyframes extracted from 92 video sequences and the testing set contains 1111 keyframes extracted from 11 video sequences. Scenes with occluded objects and different lighting-conditions are split randomly across the training/testing sets. In addition, 20000 synthetic images are provided for training on our dataset by randomly placing objects in a scene as described in section \ref{sec:synthetic-data}. Users of DTTD can opt to render additional synthetic data using the data synthesizer provided by DTTD.

\textbf{Evaluation Metrics.} Following \cite{xiang2018posecnn}, we can evaluate baseline methods with the average distance metrics ADD and ADD-S. Suppose $R$ and $T$ are ground truth rotation and translation and $\tilde{R}$ and $\tilde{T}$ are predicted rotation and translation. The ADD metric computes the mean of the pairwise
distances between the 3D model points using ground truth pose $(R, T)$ and predicted pose $(\tilde{R}, \tilde{T})$:
$$\mathrm{ADD} = \frac{1}{m}\sum_{x \in M} \| (Rx + T) - (\tilde{R}x + \tilde{T})\|,$$
where $M$ denotes the point set sampled from the object's 3D model and $x$ denotes the point sampled from $M$. 

The ADD-S metric is designed for symmetric objects when the matching between points could be ambiguous. The formula is expressed through the closest point distance between $x_1, x_2 \in M$:
$$\mathrm{ADD{-}S} = \frac{1}{m}\sum_{x_1 \in M}\min_{x_2 \in M} \| (Rx_1 + T) - (\tilde{R}x_2 + \tilde{T})\|$$



Following \cite{xiang2018posecnn, wang2019densefusion, liu2021stereobj1m}, a 3D pose estimate can be considered to be correct if the average distance error is smaller than a pre-defined threshold. To avoid calculating the success rate only with respect to an ad hoc threshold, we measure AUC (i.e., the area under the success-threshold curve) under various distance thresholds, with the threshold value having the x-axis and can be normalized to a relative range between 0 and 1. As a result, AUC with its values also between 0 and 1 can be calculated as a performance metric of ADD and ADD-S.


\subsection{Baseline Methods}\label{baseline-methods}
We evaluate DenseFusion \cite{wang2019densefusion} and DenseFusion with abc loss \cite{Jiang_2022_CVPR}. We also provide evaluations on FFB6D \cite{He_2021_CVPR} with and without hole filling algorithm \cite{ku2018defense} applied. We train the above methods on the training sets with synthetic data and report their performance on the test set.

\textbf{DenseFusion}\cite{wang2019densefusion} is a holistic method that directly regresses 3D pose of objects in a given RGB-D image. The model extracts features from RGB images and point clouds using separate CNN and point
cloud encoders and then performs per-pixel dense fusion with confidence score for pose estimation.

\textbf{DenseFusion with abc loss\cite{Jiang_2022_CVPR}.} Jiang et al. \cite{Jiang_2022_CVPR}'s work demonstrate the benefits of abc loss on improving model's performance on ADD AUC metric. The abc loss $\mathcal{L}_{\text{abc}}$ is defined as the L1 distance between the points sampled on the objects model in the ground truth pose and corresponding points on the same model transformed by the estimated pose.

\textbf{FFB6D}\cite{He_2021_CVPR} is a keypoint-based method that recovers object pose parameters by regressing on the most confident 3D keypoint features given an RGB-D image. The model extracts the pointwise RGB-D features for 3D keypoints localization of each object using a full flow bidirectional fusion network that facilitates the learning of an object's geometric and appearance representation. A Least-Squares Fitting algorithm is applied on selected 3D keypoints in the object coordinate system and camera coordinated system to obtain optimized pose parameters. Note that the hole filling algorithm \cite{ku2018defense} becomes a design choice when dealing with sparse depth data.

\begin{table*}
    \centering
    \resizebox{\textwidth}{!}{
    \begin{tabular}{|c|c|c|c|c|c|c|c|c|c|c|c|}
    \hline
         & \multicolumn{2}{|c|}{DenseFusion\cite{wang2019densefusion} (per-pixel)} & \multicolumn{2}{|c|}{DenseFusion\cite{wang2019densefusion} (per-pixel) + abc loss\cite{Jiang_2022_CVPR}} 
         & \multicolumn{2}{|c|}{FFB6D\cite{He_2021_CVPR}} & \multicolumn{2}{|c|}{FFB6D\cite{He_2021_CVPR} without hole filling\cite{ku2018defense}} \\
    \hline
         Object & ADD-S AUC & ADD AUC & ADD-S AUC  & ADD AUC & ADD-S AUC & ADD AUC & ADD-S AUC & ADD AUC \\
    \hline
         \text{mac\_cheese} & 84.8607 & 59.4370 & 84.4954 & 62.3326 & 48.2619 & 18.9750 & 32.4646 & 14.2986 \\
    \hline
         \text{spam\_can} & 87.9855 & 38.8337 & 88.3336 & 51.7568 & 30.6463 & 8.7439 & 46.1623 & 15.3382 \\
    \hline
         \text{mustard} & 84.3483 & 57.6207 & 84.4658 & 59.2276 & 46.6096 & 28.5681 & 24.4950 & 16.8288 \\
    \hline
         \text{pink\_expo\_marker} & 89.9759 & 78.9417 & 92.3427 & 80.7950 & 47.1057 & 39.8565 & 35.8469 & 30.5378 \\
    \hline
         \text{cheerios\_box} & 71.3897 & 29.1047 &
         69.1718 & 32.8976 & 35.4673 & 7.0891 &
         41.6812 & 7.7160 \\
    \hline
         \text{black\_expo\_marker} & 79.2332 & 62.8041 & 80.9540 & 62.6692 & 6.6660 & 5.6904 & 0.7048 & 0.6436 \\
    \hline
         \text{tomato\_can} & 84.9898 & 38.6220 & 87.1845 & 45.5703 & 24.3613 & 13.8302 & 7.7058 & 2.8534 \\
    \hline
         \text{cheez-it\_box} & 86.2357 & 60.0661 & 85.4747 & 49.1522 & 29.6672 & 7.9421 & 31.9884 & 13.9151 \\
    \hline
         \text{clam\_can} & 93.2859 & 71.6521 & 95.5851 & 76.8125 & 24.1573 & 9.6682 & 24.0257 & 12.8887 \\
    \hline
         \text{pop-tarts\_box} & 79.8134 & 17.4611 & 82.3599 & 24.1388 & 45.9755 & 2.9503 & 21.9485 & 0.5714 \\
    \hline
         \text{Average} & 84.2118 & 51.4543 & 85.0368 & 54.5353 & 37.4527 & 14.3314 &  26.7023 & 11.5592 \\
    \hline
    \end{tabular}}
    \caption{AUC results of ADD-S and ADD on the DTTD dataset. }
    \label{tab:eval_dttd}
\end{table*}

\subsection{Experimental Results and Dataset Analysis} \label{baseline-results}
\textbf{Implementation Detail.} We train the DenseFusion and DenseFusion with abc loss (we set $\alpha=1$ in overall learning objective described in section \ref{baseline-methods}) with batch size of 1 for 64 epochs. Notice that we train and evaluate both methods under given groundtruth segmentation of objects without post refinement process. We also apply ImageNet \cite{5206848} pretrained ResNet-18 \cite{He_2016_CVPR} as the CNN encoder for both methods. For FFB6D, we follow the proposed SIFT-FPS algorithm\cite{He_2021_CVPR} to obtain 8 target 3D keypoints for each object in DTTD. FFB6D also applies a hole filling algorithm \cite{ku2018defense} as an expensive postprocessing step on depth data. In our experiments, we train FFB6D with a batch size of 1 for 10 epochs for best checkpoint. We also choose to train FFB6D without applying the hole filling algorithm with a batch size of 1 for 7 epochs for best checkpoint. The performance of FFB6D with and without applying hole filling algorithm is then compared. Since object segmentation is trained together with FFB6D, we do not apply groundtruth mask like we did on training DenseFusion. All methods are adjusted to input image resolution of $1280 \times 720$. All experiments are conducted with one GTX 1080-Ti GPU.

\begin{figure}
        \centering
        \includegraphics[width=1.0\linewidth]{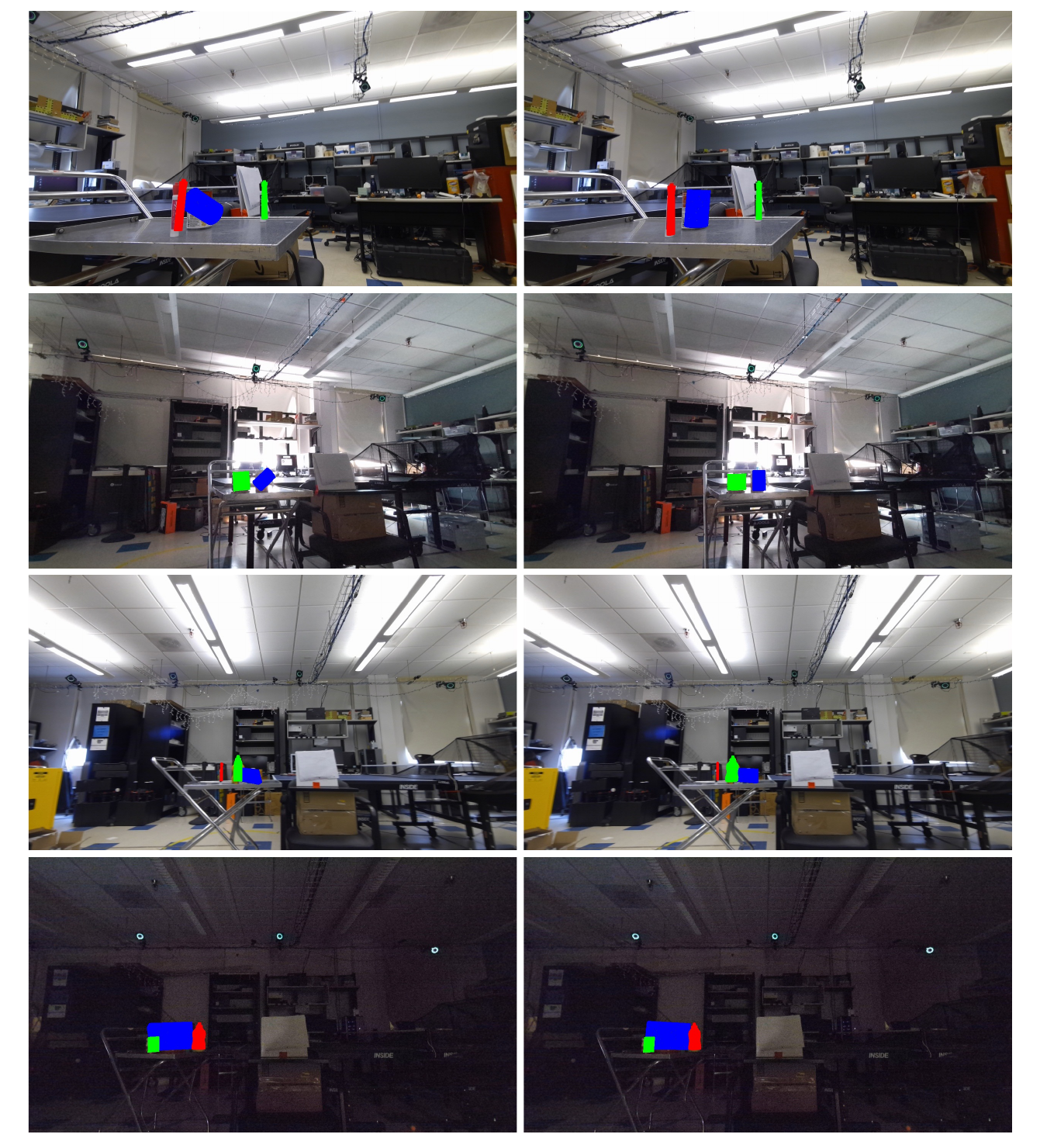}
    \caption{\textbf{Visual comparison of DenseFusion.} The left and right columns show results of without and with abc loss, respectively.}
    \label{fig:visualization}
\end{figure}

\begin{figure}
        \centering
        \includegraphics[width=1.0\linewidth]{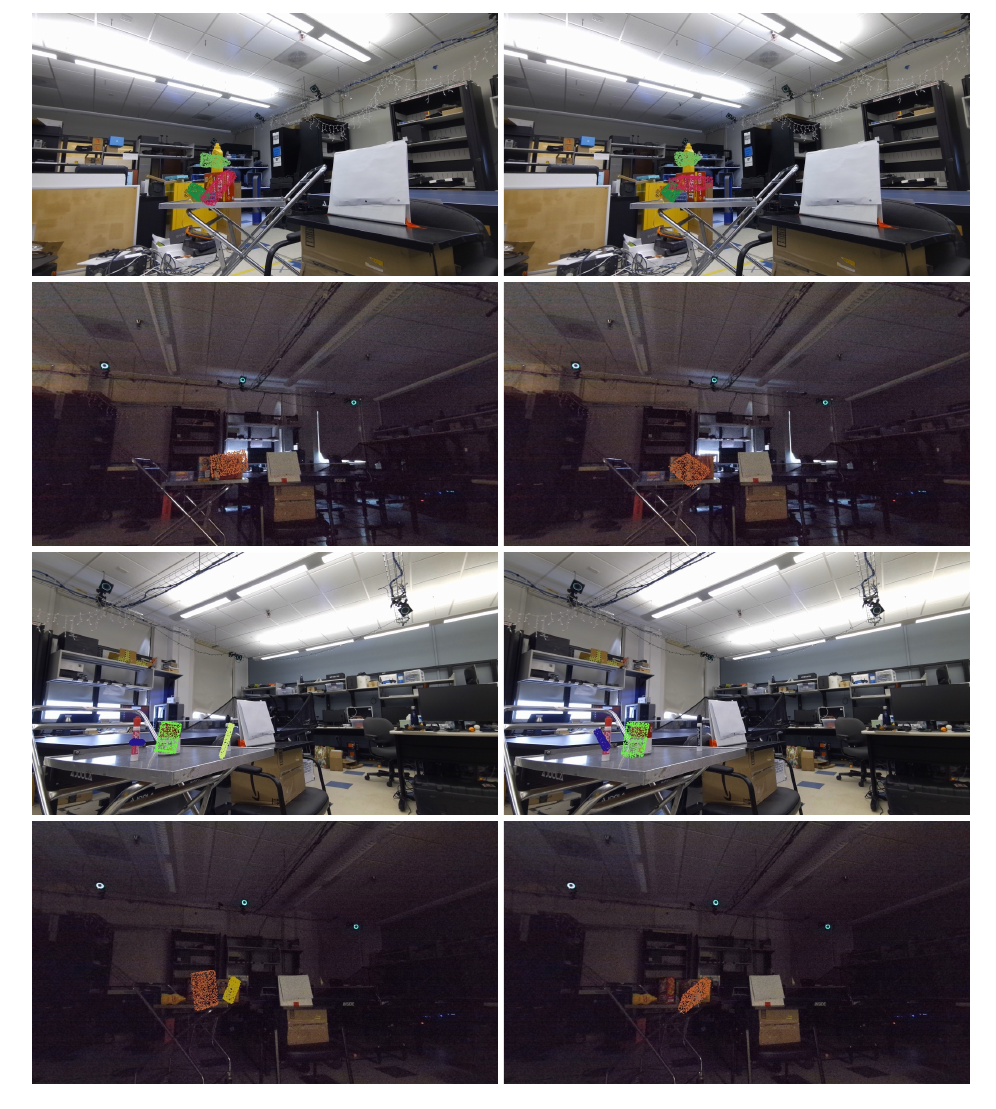}
    \caption{\textbf{Illustration of FFB6D results.} The left and right columns show the evaluation of FFB6D with and without the hole filling algorithm applied, respectively. Both methods fail to effectively segment objects in dark lighting conditions.}
    \label{fig:ffb6d_visualization}
\end{figure}

\textbf{Baseline Evaluation on DTTD Dataset.} Quantitative results on the DTTD dataset
are shown in Table \ref{tab:eval_dttd}. DenseFusion achieves average score of 84.21 ADD-S AUC and 51.45 ADD AUC. DenseFusion with abc loss has better performance over DenseFusion on most of the objects. Some qualitative evaluation results are shown in Figure \ref{fig:visualization}. FFB6D achieves average score of 37.4527 ADD-S AUC and 14.3314 ADD AUC. FFB6D without hole filling has average scores of 26.7023 ADD-S AUC and 11.5592 ADD AUC. Given that we do not apply groundtruth mask to train both methods in FFB6D, we can observe some failure cases in FFB6D's segmentation network as shown in Figure \ref{fig:ffb6d_visualization}. We suspect that FFB6D may have performed incorrect key-point matching due to the complex background in scenes. Also, we observe that dark lighting conditions can make the FFB6D segmentation network harder to learn the detection of objects in a longer range. 

\textbf{Effect of abc loss on DenseFusion.} As shown in Table \ref{tab:eval_dttd}, DenseFusion with abc loss trained on DTTD dataset outperforms DenseFusion without abc loss by 3.0\% of ADD improvements. From an optimization standpoint, the difference in abc loss and RT regression loss is the difference in L1-norm and L2-norm. Numerically, L2-norm squares value of difference in distance, while L1-norm only takes the absolute value of such. L2-norm is more outlier-prone since it increases the cost of outliers exponentially whereas the L1 norm considers them linearly. Thus, adding L1 regularization loss during training improves model robustness towards outlier, which is especially important for the ADD performance in 6-DoF pose estimation task on non-symmetric objects. Therefore, adding an L1 regularization loss helps jump out of the local minima, especially in non-symmetric objects cases.

\textbf{Effect of hole filling algorithm on FFB6D.} We notice that the performance of FFB6D with or without hole filling is unstable in DTTD dataset. Since the purpose of FFB6D employing depth hole-filling techniques is to deal with stereo depth holes issues with some of data in YCB-Video. However, the hole filling algorithm itself should not affect the model's performance if the depth data does not suffer from above issues, such as in DTTD using Azure Kinect camera. Besides, at a longer distance, the hole filling algorithm might induce a larger interpolation error. Thus we observe a different performance in different objects and scenes.

\textbf{Performance Comparison with YCB-Video Dataset.} With ground-truth segmentation mask provided, the performance of DenseFusion on DTTD does not have a comparable ADD-S AUC and ADD AUC performance against that of DenseFusion on YCB-Video dataset, as shown in Table \ref{tab:ycb_dttd_comparison}. We stipulate that DTTD poses the following additional challenges to the 6-DoF pose estimation task \footnote{Object correspondence details (left: DTTD objects; right: YCB-Video objects): \text{cheez-it\_box} with \text{003\_cracker\_box}, \text{tomato\_can} with \text{005\_tomato\_soup\_can}, \text{mustard} with \text{006\_mustard\_bottle}, \text{spam\_can} with \text{010\_potted\_meat\_can}, \text{black\_expo\_marker} with \text{040\_large\_marker}}. 
\begin{itemize}
\item Firstly, compared to YCB-Video, objects captured in DTTD include longer distances on average with wider field of view. Therefore, it becomes challenging for the model to learn the object's low-level features, such as geometrical and appearance information. 
\item Secondly, DTTD provides more lighting variations than YCB-Video. The model also needs to bridge the gap of tracking 3D objects under varying lighting conditions, e.g., estimate \text{black\_expo\_marker}'s pose under dark lighting conditions. 
\end{itemize}
\begin{table}
    \centering
    \resizebox{0.475 \textwidth}{!}{
    \begin{tabular}{|c|c|c|c|c|}
    \hline
         & \multicolumn{2}{|c|}{DTTD} & \multicolumn{2}{|c|}{YCB-Video \cite{xiang2018posecnn}} \\
    \hline
         Object & ADD-S AUC & ADD AUC & ADD-S AUC & ADD AUC \\
    \hline
         \text{spam\_can} & 87.9855 & 38.8337 & 90.3031 & 79.9354 \\
    \hline
         \text{mustard} & 84.3483 & 57.6207 & 97.0297 & 94.9152 \\
    \hline
         \text{black\_expo\_marker} & 79.2332 & 62.8041 & 97.2883 & 93.9526 \\
    \hline
         \text{tomato\_can} & 84.9898 & 38.6220 & 96.6725 & 87.7622 \\
    \hline
         \text{cheez-it\_box} & 86.2357 & 60.0661 & 93.1328 & 88.3662 \\
    \hline
    \end{tabular}}
    \caption{DenseFusion (per-pixel) performance on DTTD and YCB-Video dataset. We select overlapping objects with similar appearances in both datasets.}
    \label{tab:ycb_dttd_comparison}
\end{table}

We believe the above performance differences make DTTD a better and more up-to-date dataset to demonstrate the technical gaps in digital-twin tracking and to challenge the improvement of future digital-twin tracking solutions.

\section{Conclusion}
In this paper, we have introduced a novel 3D object tracking RGB-D dataset, called Digital Twin Tracking Dataset (DTTD), to enable further research on extending potential solutions to longer-range object tracking problems. To provide better point cloud data with less 3D noise and in longer ranges, DTTD adopts Microsoft Azure Kinect camera as the ToF depth camera sensor. The DTTD open-source code also provides a data collection pipeline that aims to minimize the ground-truth labeling error by leveraging an OptiTrack system, an ARUCO marker, and manual annotation. In total, DTTD contains 103 scenes of 10 common off-the-shelf textured objects, with each frame annotated with a per-pixel semantic segmentation and ground-truth object poses. Our experiments have evaluated different methods using DTTD and have demonstrated that DTTD can help evaluate object tracking methods under common digital-twin conditions and illustrate new challenges in longer-range and wider lighting conditions.

For future work, we plan to further update the DTTD dataset with more objects, scenes, backgrounds, and even newer RGB-D camera sensors that are relevant to AR applications.

\section*{Acknowledgement}
This work is supported by the FHL Vive Center at the University of California, Berkeley. We would like to acknowledge Tianjian Xu, Han Cui, Yunhao Liu, Zixun Huang, and Keling Yao for assisting with the DTTD data collection process.

{\small
\bibliographystyle{ieee_fullname}
\bibliography{egbib}
}

\end{document}